\documentclass[journal]{IEEEtran}
\ifCLASSINFOpdf
\else
\fi

\usepackage{multirow}
\usepackage{algorithm}
\usepackage{algpseudocode}
\usepackage{color}
\usepackage{graphicx}
\usepackage{amsmath}
\usepackage{amssymb}
\newtheorem{definition}{Definition}

\begin{document}
%
\title{Line Graph Neural Networks for Link Prediction}
%
%
%

\author{Lei Cai, Jundong Li, Jie Wang, Shuiwang Ji ~\IEEEmembership{Senior~Member,~IEEE}
	\thanks{Lei Cai is with the School of Electrical Engineering and Computer Science, Washington State University, Pullman, 
		WA 99164. E-mail: lei.cai@wsu.edu.}
	\thanks{Shuiwang Ji is with the Department of Computer Science and Engineering, Texas A\&M University,
		College Station, TX 77843. E-mail: sji@tamu.edu.}
	\thanks{Jundong Li is with the Department of Electrical and Computer Engineering, Department of Computer Science, and School of Data Science, University of Virginia, Charlottesville, VA 22904. E-mail: jundong@virginia.edu.}
	\thanks{Jie Wang is Department of Electronic Engineering and Information Science, University of Science and Technology of China, China 230026. Email: jiewangx@ustc.edu.cn}}

%
%

\markboth{IEEE Transactions on Pattern Analysis and Machine Intelligence}%
{Cai \MakeLowercase{\textit{et al.}}: }
%



\maketitle

\begin{abstract}
    We consider the graph link prediction task, which is a classic graph analytical problem with many real-world applications. With the advances of deep learning, current link prediction methods commonly compute features from subgraphs centered at two neighboring nodes and use the features to predict the label of the link between these two nodes. In this formalism, a link prediction problem is converted to a graph classification task. In order to extract fixed-size features for classification, graph pooling layers are necessary in the deep learning model, thereby incurring information loss. To overcome this key limitation, we propose to seek a radically different and novel path by making use of the line graphs in graph theory. In particular, each node in a line graph corresponds to a unique edge in the original graph. Therefore, link prediction problems in the original graph can be equivalently solved as a node classification problem in its corresponding line graph, instead of a graph classification task. Experimental results on fourteen datasets from different applications demonstrate that our proposed method consistently outperforms the state-of-the-art methods, while it has fewer parameters and high training efficiency.
\end{abstract}


%
\IEEEpeerreviewmaketitle
\IEEEPARstart{L}{ink} prediction models are used to learn the distribution of links
in graphs and predict the existence of potential links
\cite{liben2007link,schlichtkrull2018modeling,liben2007link,al2006link,lu2011link}.
In many real-world applications, the input data is represented as
graphs, and link prediction models can be applied to tasks like
friend recommendation in social networks \cite{adamic2003friends},
product recommendation in e-commerce
\cite{koren2009matrix,wang2019knowledge,you2019hierarchical},
knowledge graph completion \cite{nickel2015review}, protein
interaction analysis \cite{airoldi2008mixed}, and metabolic network
reconstruction \cite{oyetunde2017boostgapfill}.

To solve the link prediction problem, various heuristic methods were proposed
to measure the similarity between two target nodes and predict the
existence of link \cite{barabasi1999emergence}. However, 
the heuristic functions in these methods are often manually
designed for a specific network, limiting their applicability to
diverse areas. For example, the number of common neighbors \cite{adamic2003friends} is
employed as a first-order heuristic function to predict the
potential friendship relations in social networks and achieves satisfactory
performance. However, this heuristic may not work well on
protein-protein interaction networks, since two proteins sharing
many common neighbors may have a low probability of interacting
\cite{kovacs2019network}.

Many heuristic methods have been proposed to solve graph link
prediction problems from different areas. 
However, there still exist challenges to select heuristic functions given a new network. To tackle these
challenges, the link prediction model based on graph neural
networks (SEAL) was proposed to learn heuristic functions from
$h-$hop neighborhood automatically \cite{zhang2018link}. This method
achieved the state-of-the-art performance on a variety of graphs.
The SEAL model extracts an $h$-hop enclosing subgraph centered on
the two target nodes and predicts the existence of link based on the
topology of enclosing subgraph. Therefore, the link prediction task is converted to 
the graph classification problem, where the model takes
the enclosing subgraph as inputs and predicts the existence of link
between them. Normally, graph neural networks \cite{zhang2018end} are employed
to learn features to represent the topology of the subgraph.
Therefore, graph pooling layers
\cite{gao2019graph,Yuan2020StructPool:,ying2018hierarchical} are
required to compute a fixed-size feature vector from the whole graph while some information
may be lost in this operation. For example, in sort pooling operations \cite{zhang2018end}, only partial nodes can be selected to represent the graph.
In addition, a graph neural network with pooling layers often requires more training time to
converge.

\begin{figure*}[t]
	\center
	\includegraphics[width=\textwidth]{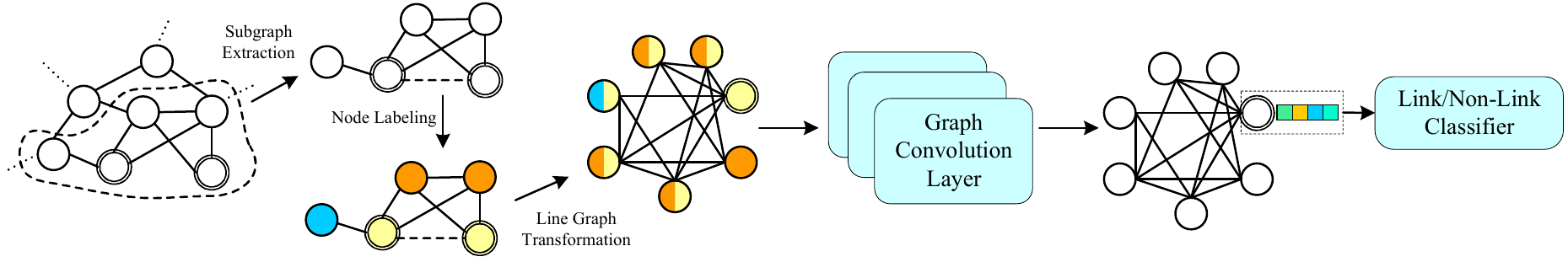}
	\caption{Illustration of our proposed model based on line graph
		neural networks. The two target nodes in the graph are marked with
		double circles. To predict the existence of the link, an $h$-hop
		enclosing subgraph centered on two target nodes is extracted. A node
		labeling function is employed to assign the label for each node to
		represent the structural importance to the target link. To learn the
		feature of the target link, we transform the enclosing subgraph into
		a corresponding line graph. The graph convolution networks are used
		to learn the feature that is employed to predict the existence of
		link.} \label{fig:pipeline}
\end{figure*}

Although the SEAL works well in many types of graphs, it still has
some limitations, due to the usage of pooling operations in the graph neural
network. To solve the information loss in pooling layers, we propose
to learn the features of the target link directly instead of
extracting features from the whole enclosing subgraph. Compared with extracting features from the whole graph,  learning node embedding is more effective. Graph
convolution layers \cite{gao2018large,kipf2016semi,niepert2016learning,hamilton2017inductive}
have shown promising performance for learning node embeddings.
However, graph convolution layers are not effective enough to learn
edge embeddings from graphs. To address this issue, we propose to
convert the original enclosing subgraph into a corresponding line
graph. Each node in the line graph has a unique corresponding edge
in the original graph. In addition, the topology information can be
well preserved during the transformation. Therefore, graph
convolution layers can be directly applied to learn the node embeddings in the
line graph. The node embeddings in the line graph are used as features
for the edges in the original graph to predict the existence of 
links. Therefore, the link prediction task can be regarded as the node
classification problem in our proposed framework. Our contributions can be summarized as follows:
\begin{enumerate}
	\item We analyze the limitations of exiting deep learning based link prediction methods due to graph pooling operations and envision the need to develop a new method that do not have the information loss problem.
	\item We propose a line graph neural networks model for the link prediction task, where the original graph is transformed into a corresponding line graph to enable efficient feature learning for the target link.
	\item We conduct experiments on 14 datasets from different areas. Our proposed method can achieve promising performance on different datasets and outperform all baseline methods, including the previous state-of-the-art model.
	\item Our proposed method can achieve promising performance only with graph convolution layers, and thus requires fewer parameters. In addition, the neural network consisting of graph convolution layers converges significantly faster than the state-of-the-art model.
\end{enumerate}

\section{Related Work}

Link prediction models can be grouped into three categories -- heuristic methods, embedding methods, and deep learning methods. 

\textbf{Heuristic Methods:} The key idea of heuristic methods is to compute the similarity score from the neighborhood of two target nodes. Based on the maximum hop of neighbors used in the computation procedure, heuristic methods can be categorized into three groups, including first-order, second-order, and high-order heuristics. Common neighbors and preferential attachment \cite{barabasi1999emergence} are typical first-order heuristics since only one-hop neighbors are employed to compute the similarity.  Second-order heuristic methods that involve two-hop neighbors include Adamic-Adar \cite{adamic2003friends} and resource allocation \cite{barabasi1999emergence, zhou2009predicting}. In addition, high-order heuristics, including Katz \cite{katz1953new}, rooted PageRank \cite{brin2012reprint}, and SimRank \cite{jeh2002simrank} were proposed to compute the similarity score between a pair of nodes using the whole graph. High-order heuristic methods can often achieve better performance than low-order heuristics but require more computation cost. Since many heuristic methods were proposed to handle different graphs, selecting a favorable heuristic method becomes a challenging problem. 

\textbf{Embedding Methods:} The similarity between two target nodes can also be calculated based on node embeddings \cite{ng2002spectral}. Therefore, embedding methods that can learn the features of nodes from graph topology were also employed to solve the link prediction task, and typical methods along this line include matrix factorization \cite{koren2009matrix} and stochastic block \cite{airoldi2008mixed} etc. Recently, inspired by world embedding methods in natural language processing tasks, recent advances such as deepwalk \cite{perozzi2014deepwalk}, LINE \cite{tang2015line}, and node2vec \cite{qiu2018network} were proposed to learn node embedding via the skip-gram method. Deepwalk generates random walks for each vertex with a given length and picks the next visited node uniformly from the neighbors of the current node. Later on, the skip-gram method is employed to learn node embeddings from the generated node sequence. The node embedding methods can learn informative features from the graph and thus achieve satisfactory performance for the link prediction task. However, the performance of link node embedding methods can be affected if the graph becomes very sparse.

\textbf{Deep Learning:} To overcome the limitations of heuristic methods, deep learning based methods were proposed to learn the distribution of links from the graph automatically \cite{zhang2017weisfeiler, zhang2018link, cai2020link}. Weisfeiler-Lehman Neural Machine was proposed to predict the existence of a link using a fully-connected neural network based on a fixed-size enclosing subgraph centered on the two target nodes \cite{zhang2017weisfeiler}. To predict the existence of a link from a general enclosing subgraph, SEAL \cite{zhang2018link} converts the link prediction task to a graph classification problem and solve it using graph neural networks. Due to the promising learning ability of graph neural networks, the SEAL model achieves the state-of-the-art performance for the link prediction problem. Later on, a multi-scale link model was also proposed to extend SEAL to achieve better performance on plain graphs \cite{cai2020link}.

\section{The Proposed Methods}

\subsection{Problem Formulation}

In the link prediction task, we are often given a network represented as an undirected graph $G=(V,E)$ which consists of a set of vertices $V=\{v_1, v_2, ..., v_n\}$ and a set of links $E \subseteq V \times V$. The graph can also be represented by the adjacency matrix $A$. If there exists a link between vertex $i$ and $j$, then $A_{i,j}=1$ and $A_{i,j}=0$ otherwise. The goal of link prediction is to predict potential or missing links that may appear in a foreseeable future. 

\subsection{Overall Framework}

Deep learning based link prediction models were proposed to learn the link distribution from the existing links and determine whether a link exists between two target nodes in the graph. For example, when we predict if there exists a link between two users in a social network, the number of mutual friends is commonly considered as a main criterion. If two users share many mutual friends, they are more likely to be connected. In this sense, if the $1$-hop subgraph induced from two target nodes are densely connected, we will have higher chances to observe a link between them. Considering the variation of networks from different areas, deep learning based methods were proposed to learn the topology feature of subgraphs automatically and predict the existence of links\cite{zhang2018link}. In particular, the deep learning based link prediction models generally consist of the following three components:	
\begin{enumerate}
	\item Enclosing subgraph extraction: The existence of a potential link can be determined by the topology of a local enclosing subgraph centered on two target nodes. To seek a balance between computation cost and prediction performance, an $h-$hop enclosing subgraph is extracted for learning features and predicting the existence of potential links. 
	\item Node labeling: Given an enclosing subgraph, we are required to identify the role of each node in the graph before learning features and predict the existence of the link. That is, we need to identify the target nodes and mark the structural importance of other nodes. A favorable labeling function is of great importance for the further feature learning procedure.
	\item  Feature learning and link prediction: The output of node labeling function can be used as the attribute of each node in the graph. The attribute can indicate the structural importance of the link to be predicted. Graph neural networks are commonly employed to learn features from the given enclosing subgraph, which can be further used to predict the existence of a link.
\end{enumerate}

In this work, we propose line graph neural networks for the link prediction task. Our proposed model can be illustrated in Figure \ref{fig:pipeline}. Following the general framework of deep learning based link prediction models, we extract an $h$-hop enclosing subgraph centered on two target nodes and assign each node with a label that can represent the structural importance to the target link. The key contribution of our proposed method is the feature learning component. In the previous state-of-the-art model, graph convolution and graph pooling layers are employed to obtain a fixed-size feature vector to predict the existence of the link considering the scale variation of different graphs. Since this graph pooling layer is employed in the state-of-the-art model, only part of graph information can be preserved for further prediction. To overcome the limitations of the SEAL method, we propose to convert the enclosing subgraph to a line graph where each node corresponds to a unique link in the original graph. The feature of the link can be learned directly using the entire input from line graph representation. Thus the proposed method can greatly improve the performance of the link prediction.

\subsection{Line Graph Neural Networks}

\textbf{Line Graph Space Transformation} In order to predict the existence of a link, graph neural networks are employed to learn features from a given enclosing subgraph $G_{v_1,v_2}^h$, where $G_{v_1,v_2}^h$ is an $h$-hop enclosing subgraph centered on two target nodes $v_1$ and $v_2$, and each node in the enclosing subgraph is associated with a label that can indicate the structural importance to the target link.  Different enclosing subgraphs commonly contain a different number of nodes. To extract a fixed-size feature vector for the further prediction, we will lose some information during the procedure. To overcome this challenge, we propose to convert the enclosing subgraph to the line graph, which represents the adjacencies between edges of the original graph. Thus, the feature of the link to be predicted can be learned directly in the line graph representation using graph convolution neural networks. 

\begin{figure}[t]
	\center
	\includegraphics[width=\columnwidth]{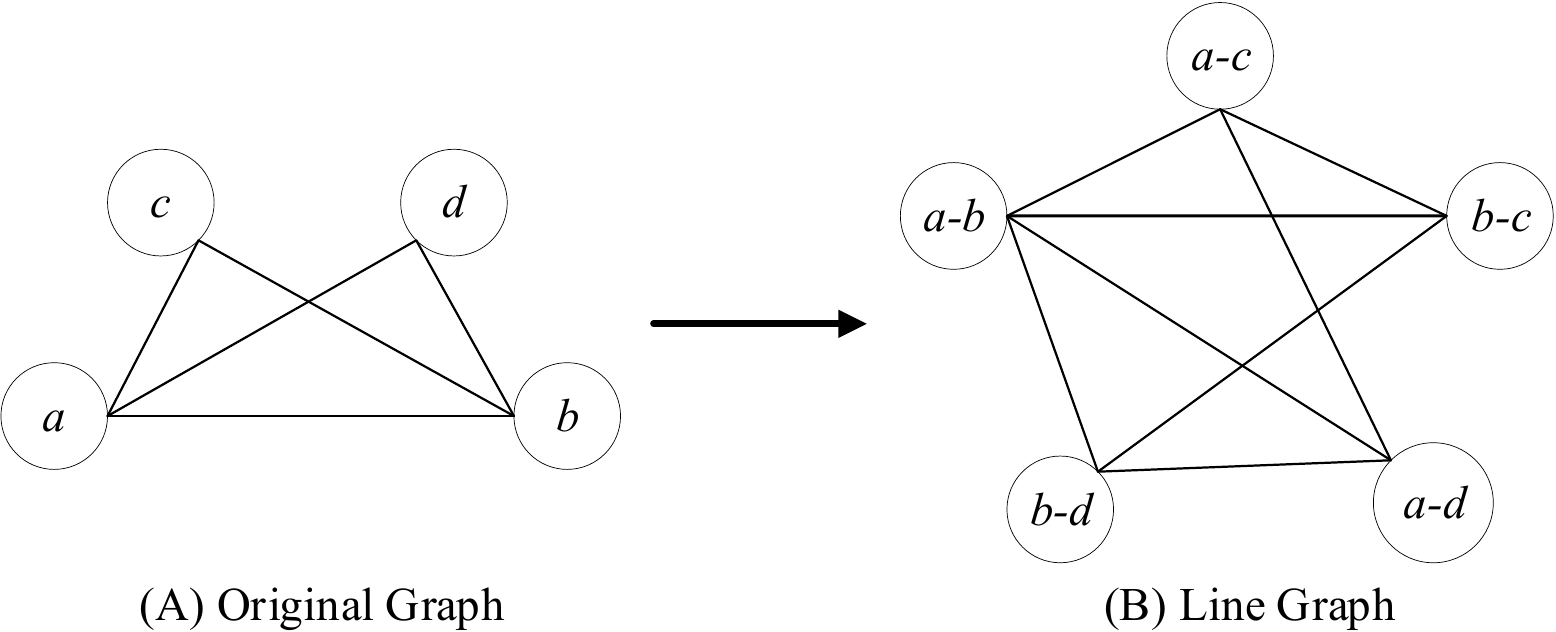}
	\caption{Illustration of the line graph transformation procedure. Each node in the line graph corresponds to a unique edge in the original graph and is marked with the name of two end nodes.
	} \label{fig:line}
\end{figure}

The line graph $L(G)$ of a given undirected graph $G$ is proposed to represent the adjacencies between edges of $G$ \cite{harary1960some, chen2018supervised}. The definition of line graph $L(G)$ can be defined as follows. 
\begin{definition}
	The edges in the original graph $G$ are considered as nodes in the line graph $L(G)$. Two nodes in $L(G)$ are connected if and only if the two corresponding links share the same node. 
\end{definition}
An example of the line graph transformation procedure is illustrated in Figure \ref{fig:line}. The original undirected graph $G$ contains four nodes and five edges. Therefore, the line graph $L(G)$ contains five nodes. The node $(a-b)$ and $(a-c)$ in the line graph are connected since the corresponding edges in the original graph $G$ share a common node $a$ based on the definition of the line graph. 

Based on the definition of the line graph, we can obtain the following property of $L(G)$: \emph{Given a graph $G$ with $m$ nodes and $n$ edges, the number of nodes of the line graph $L(G)$ equals to $n$. The number of edges in $L(G)$ is $\frac{1}{2}\sum_{i=1}^{m}d_i^2-n$, where $d_i$ is the degree of node $i$ in graph $G$.}

This property guarantees that learning features in the line graph space will not increase the computation complexity significantly. In additional, converting a graph $G$ to line graph $L(G)$ only costs linear time complexity \cite{roussopoulos1973max,lehot1974optimal}.

\textbf{Node Label Transformation} An enclosing subgraph can be converted into the corresponding line graph through transformation. However, this procedure can only transform the topology of a given graph. Each node in the enclosing subgraph also contains a label $l\in \mathbb{R}$ generated by labeling function as node attributes that can represent the structural importance. During the transformation procedure, edges in the original graph are represented as nodes in the line graph. The label $l$ is only assigned for nodes in the original graph. To transfer the node label from the original graph directly, a transformation function is required to convert the node label to edge attribute. Thus the edge attribute can be assigned directly as the node attribute in the line graph. In this work, we propose to generate the edge attribute from the node label through the following function:
\begin{equation} \label{eq: concate} 
	l_{(v_1, v_2)} = \textrm{concate}(\min(f_l(v_1),f_l(v_2)), \max(f_l(v_1),f_l(v_2))),
\end{equation}
where $f_l(\cdot)$ is the node labeling function, $v_1$ and $v_2$ are the two end nodes of the edge, and $\textrm{concate}(\cdot)$ represents the concatenation operation for the two inputs. 
Since we only consider undirected graph link prediction in this work, the attribute of edge $(v1, v2)$ and $(v2, v1)$ should be the same. It is easy to prove that the edge attribute generated by equation (\ref{eq: concate}) is consistent when switching the end nodes. In addition, the structural importance information of the node can be well preserved in the function.

The proposed method in equation (\ref{eq: concate}) can well address the edge attribute transformation in plain graphs. In some cases, graphs are commonly provided with node attributes. For example, in citation networks, the node attribute describing a summary of the paper can be provided in the graph. For attributed graphs, node attributes also play an important role in the link prediction task. Therefore, the edge attribute transformation function should be generalized to deal with attributed graph. Following the edge attribute transformation function in equation (\ref{eq: concate}), we can concatenate the original node attribute with the node label as the edge attribute. But the edge attribute will not be consistent when we switch the order of two end nodes in the undirected graph. To overcome this limitation, we propose to deal with the original node attribute and node label in different ways by:
\begin{equation} \label{eq: concate_att} 
	l_{(v_1, v_2)} = \textrm{concate}(\min(f_l(v_1),f_l(v_2)), \max(f_l(v_1),f_l(v_2)), X_{v_1}+X_{v_2}),
\end{equation}
where $X_{v_1}$ and $X_{v_2}$ are the original attribute of node $v_1$ and $v_2$.  We propose to combine the node attribute using summation operation, which can guarantee the invariance of edge attributes when switching the end nodes. The generated edge attribute $l_{(v_1, v2)}$ can be used as the node attribute directly in the line graph. Therefore, the link prediction task is converted to a node classification problem which can be solved by graph convolution neural networks.

\textbf{Feature Learning by Graph Neural Networks} With recent progress in graph neural networks, learning the graph feature becomes a favorable solution that has been explored in various graph analytical tasks \cite{Yuan2020StructPool:, kipf2016semi, ying2018hierarchical}. In this work, we employ graph convolution neural networks to learn the node embedding in the line graph, which can represent an edge in the original graph. Thus, the node embedding in the line graph can be used to predict whether a potential link is likely to exist in the network.

Given a line graph representation of the enclosing subgraph $L(G_{v1,v2}^h)$, the node embedding of $(v_i, v_j)$ in the $k$-th layer of the graph convolution neural network is indicated as $Z_{(v_i,  v_j)}^{(k)}$. Then the embedding of $(v_i, v_j)$ in the $(k+1)$-th layer is given by:
\begin{equation}\label{eq:gcn}
	Z_{(v_i, v_j)}^{(k+1)} = (Z_{(v_i, v_j)}^{(k)} + \beta \sum_{d \in \mathcal{N}_{(v_i, v_j)}} Z_d^{(k)})W^{(k)}, 
\end{equation}
where $\mathcal{N}_{(v_i, v_j)}$ is the set of neighbors of node $(v_i, v_j)$ in the line graph, $W^{(k)}$ is the weight matrix for the $k$-th layer, $\beta$ is a normalization coefficient. The input for the first layer of graph convolution neural network is set to node attribute in the line graph as $Z_{(v_i,  v_j)}^{0} = l_{(v1, v2)}$. We then consider the link prediction task as a binary classification problem and train the neural network by minimizing the cross-entropy loss for all potential links as:
\begin{equation}
	\mathcal{L}_{CE} = -\sum_{l \in L_t} (y_l\log(p_l) + (1-y_l)\log(1-p_l)),
\end{equation}
where $L_t$ is the set of target links to be predicted, $p_l$ is the probability that the link $l$ exists in the graph, and $y_l \in \{0,1\}$ is the label of a target link that indicating whether the link exists or not.

\textbf{Connection with Learning on Original Graphs} The key idea of our proposed method is to learn edge features from the enclosing subgraph and predict the existence of edge using the features. In this work, the feature of edge $e=(v_1, v_2)$ is learned based on attributes of two end nodes as:
\begin{equation}
	f_e = g(f_l(v1), f_l(v2)),
\end{equation}
where $f_e$ is the edge feature, $g(\cdot)$ is the graph neural network function. Although graph convolutional layers are performed on the line graph, it still has connections with the same operation on the original graph. We use the first layer graph convolution layer as an example to illustrate this relationship. We reformulate the equation (\ref{eq:gcn}) as:
\begin{eqnarray} \label{eq:lgcn}
	Z_{(v_i, v_j)}^1 = (l_{(v1, v2)} &+& \beta \sum_{d_1 \in \mathcal{N}_{v1}} \sum_{d_2 \in \mathcal{N}_{d1}} l_{(d1, d2)} \nonumber \\ 
	&+& \beta \sum_{d_3 \in \mathcal{N}_{v2}} \sum_{d_4 \in \mathcal{N}_{d3}} l_{(d3, d4)}) W^{(0)}.
\end{eqnarray}
The graph convolution operation learns embedding for each node by aggregating node embedding from its $1-$hop neighbors. It can be seen from equation (\ref{eq:lgcn}) that the graph convolution on the line graph can aggregate the node embedding from $2-$hop neighbors. In the line graph transformation procedure, each node attribute is derived from two corresponding node attributes. That is, the attribute of each node in the line graph contains attributes from two nodes in the original graph. Therefore, aggregating information from $1-hop$ neighbors is equivalent to performing the same operation on $2-$hop neighbors. It also shows that learning node embedding through graph convolution in the line graph is more efficient than that in the original graph in terms of neighbor embedding aggregation. 

\subsection{The Proposed Algorithm}
In this section, we provide a detailed description of the three components for our proposed framework. There is no strict restriction for the three steps. The given enclosing subgraph extraction and graph topology labeling function in this section can work well for most networks.

\textbf{Enclosing Subgraph Extraction} The existence of the link between two nodes can be determined by the graph topology centered on them. In general, we can achieve better performance when more topology information is involved. However, it will incur more computation cost. To seek a balance between performance and computation cost, we predict the existence of the link between node $v_i$ and $v_j$ using  $2-$hop enclosing subgraph as:
\begin{equation}
	G_{(v_i, v_j)}^2 = \{v|\min(d(v, v_i), d(v, v_j) \leq 2)\},
\end{equation}
where $d(v, v_i)$ is the shortest path$/$geodesic distance between $v$ and $v_i$.

\textbf{Node Labeling} Given an enclosing subgraph, we only know the topology of the graph. Before we learn features of the target link, we need to identify the role of each node in the graph through a labeling function. The node labeling function must satisfy the following criteria: (1) Identifying the two target nodes. (2) Provide the structural importance of each node to the target nodes. In this work, we employ an effective node labeling function proposed by \cite{zhang2018link} as:
\begin{equation}\label{eq:label}
	f_l(v) = 1+\min(d(v,v_1), d(v, v_2))+(d_{s}/2)[(d_{s}/2)+(d_{s})\%2-1],
\end{equation}
where $d_{s} = d(v, v_1) + d(v, v_2)$, $(d_s/2)$ and $(d_s\%2)$ are the integer quotient and remainder of $d$ divided by 2, respectively. In addition, the two target nodes $v_1$ and $v_2$ are assigned with label 1 as $f_l(v_1) = 1$ and $f_l(v_2) = 1$. For any node $v$ satisfying $d(v,v_1)=\infty$ or $ d(v, v_2)=\infty$, it will be assigned with label 0 as $f_l(v)=0$. The node labeling function provides a label $f_l(\cdot) \in \mathbb{R}$. In practice, the node label is represented as a one-hot vector. As discussed above, the edge is represented as an order invariant pair. The edge feature pair is represented as a concatenation of two one-hot vectors. Algorithm \ref{alg: lglp} shows the link prediction procedure using our proposed framework.

\begin{algorithm}
	\caption{Link Prediction Model Based on Line Graph Neural Networks}  \label{alg: lglp}
	\begin{algorithmic}[1]
		\State \textbf{Input}: Target link $(v_1, v_2)$, Graph $G$
		\State \textbf{Output}: Prediction result
		\State Extract $h-$hop enclosing subgraph $G_{(v_1, v_2)}^h$
		\State Apply node labeling function (\ref{eq:label}) to $G_{(v_1, v_2)}^h$
		\State Generate edge attribute using equation (\ref{eq: concate}) or (\ref{eq: concate_att}) 
		\State Transform $G_{(v_1, v_2)}^h$ to line graph $L(G_{(v_1, v_2)}^h)$
		\State Apply graph neural networks to extract node embeddings on $L(G_{(v_1, v_2)}^h)$ to predict the existence of link
	\end{algorithmic}
\end{algorithm}

\section{Experiments}

In this section, we evaluate our proposed method on 14 different datasets for the link prediction task. Two evaluation metrics, including area under the curve (AUC) and average precision (AP) are employed in this work to measure the performance of different models. The code and dataset used in this work will be available online after the paper is published.

\begin{table}[t]
	\caption{Summary of datasets used in our experiments. The number of node, link, average node degree, and graph type are provided for each dataset.}
	\label{summary}
	\centering
	\resizebox{0.49\textwidth}{!}{
		\begin{tabular}{lccccccccc}
			\hline
			Name & \#Nodes & \#Links & Degree  & Type\\
			\hline
			\hline
			BUP & 105 & 441 & 8.4 & Political Blogs\\
			C.ele & 297 & 2148 & 14.46 & Biology\\
			USAir & 332 & 2126 & 12.81 & Transportation\\
			SMG & 1024 & 4916 & 9.6 & Co-authorship\\
			EML & 1133 & 5451 & 9.62 & Shared Emails\\
			NSC & 1461 & 2742 & 3.75 & Co-authorship\\
			YST & 2284 & 6646 & 5.82 & Biology\\
			Power & 4941 & 6594 & 2.669& Power Network\\
			KHN & 3772 & 12718 & 6.74 & Co-authorship\\
			ADV & 5155 & 39285 & 15.24 & Social Network\\
			GRQ & 5241 & 14484 & 5.53 & Co-authorship\\
			LDG & 8324 & 41532 & 9.98 & Co-authorship\\
			HPD & 8756 & 32331 & 7.38 & Biology\\
			ZWL & 6651 & 54182 & 16.29 & Co-authorship\\
			\hline  
		\end{tabular}
	}
\end{table}

\subsection{Datatsets and Baseline Models}
In this work, we perform our proposed line graph link prediction (LGLP) model on 14 different datasets, including BUP, C.ele, HPD, YST, SMG, NSC, KHN, GRQ, LDG, ZWL, USAir, EML, Power, and ADV \cite{watts1998collective,newman2001structure}. To demonstrate that our proposed method can work well in different areas, 14 datasets are collected from 6 areas. In addition, graphs in different scales, including the number of nodes and links, are used in the experiments. The details of the datasets are shown in Table \ref{summary}. 

\begin{table*}[t]
	\caption{AUC comparison with baseline methods (80\% training links).}
	\label{result801}
	\centering
	\begin{tabular}{lccccccc}
		\hline
		Model & BUP & C.ele & USAir & SMG & EML & NSC & YST\\
		\hline
		Katz & $87.10(\pm2.73)$ & $84.84(\pm2.05)$ & $92.01(\pm0.88)$ & $86.09(\pm1.06)$ & $88.45(\pm0.68)$ & $98.00(\pm0.31)$ & $80.56(\pm0.78)$ \\
		PR & $90.13(\pm2.45)$ & 89.14$(\pm1.35)$ & $93.74(\pm1.01)$ & $89.13(\pm0.90)$ & $89.46(\pm0.63)$ & $98.05(\pm0.29)$ & $81.40(\pm0.75)$ \\
		SR & $85.47(\pm2.75)$ & $75.65(\pm2.24)$ & $79.21(\pm1.50)$ & $78.39(\pm1.14)$ & $86.90(\pm0.71)$ & $97.19(\pm0.48)$ & $73.93(\pm0.95)$ \\
		N2V & $80.25(\pm5.55)$ & $80.08(\pm1.52)$ & $85.40(\pm0.96)$ & $78.30(\pm1.22)$ & $83.06(\pm1.42)$ & $96.23(\pm0.95)$ & $77.07(\pm0.36)$ \\
		SEAL & $93.32(\pm0.84)$ & $87.44(\pm1.21)$ & $95.21(\pm0.77)$ & $91.53(\pm0.46)$ & $92.01(\pm0.38)$ & $99.55(\pm0.01)$ &$90.72(\pm0.25)$ \\ 
		LGLP & \textbf{95.24$(\pm0.53)$} & \textbf{90.16$(\pm0.76$)} & \textbf{97.44$(\pm0.32)$} & \textbf{92.53$(\pm0.29$)} & \textbf{92.03$(\pm0.28)$} & \textbf{99.82$(\pm0.01)$} & \textbf{91.97$(\pm0.12)$}\\
		\hline  
		\hline
		Model & Power & KHN & ADV & LDG & HPD & GRQ & ZWL\\
		\hline
		Katz & $59.59(\pm1.51)$ & $84.60(\pm0.79)$ & $92.13(\pm0.21)$ & $92.96(\pm0.19)$ & $85.47(\pm0.35)$ & $89.81(\pm0.59)$ & $96.42(\pm0.12)$ \\
		PR & $59.88(\pm1.51)$ & $88.43(\pm0.80)$ & $92.78(\pm0.18)$ & $94.46(\pm0.19)$ & $87.19(\pm0.34)$ & $89.98(\pm0.57)$ & $97.20(\pm0.12)$ \\
		SR & $70.18(\pm0.75)$ & $79.55(\pm0.90)$ & $86.18(\pm0.22)$ & $90.95(\pm0.14)$ & $81.73(\pm0.37)$ & $89.81(\pm0.58)$ & $95.97(\pm0.16)$ \\
		N2V & $70.37(\pm1.15)$ & $82.21(\pm1.19)$ & $77.70(\pm0.83)$ & $91.88(\pm0.56)$ & $79.61(\pm1.14)$ & $91.33(\pm0.53)$ & $94.38(\pm0.51)$ \\
		SEAL & $81.37(\pm0.93)$ & $92.69(\pm0.14)$ & $95.07(\pm0.13)$ & $96.44(\pm0.13)$ & $92.26(\pm0.09)$ & $97.10(\pm0.12)$ & $97.46(\pm0.02)$ \\ 
		LGLP & \textbf{82.17$(\pm0.57)$} & \textbf{93.30$(\pm0.09)$} & \textbf{95.40$(\pm0.10)$} & \textbf{96.70$(\pm0.07)$} & \textbf{92.58$(\pm0.08)$} & \textbf{97.68$(\pm0.10)$} & \textbf{97.76$(\pm0.01)$}\\
		\hline  
	\end{tabular}
\end{table*}

\begin{table*}[t]
	\caption{AP comparison with baseline methods (80\% training links).}
	\label{result802}
	\centering
	\begin{tabular}{lccccccc}
		\hline
		Model & BUP & C.ele & USAir & SMG & EML & NSC & YST\\
		\hline
		Katz & $85.94(\pm3.46)$ & $85.94(\pm3.46)$ & $93.51(\pm0.79)$ & $87.68(\pm0.90)$ & $90.54(\pm0.53)$ & $98.02(\pm0.43)$ & $85.76(\pm0.64)$ \\
		PR & $89.53(\pm3.11)$ & 87.96$(\pm1.69)$ & $94.30(\pm1.27)$ & $91.07(\pm0.59)$ & $91.01(\pm0.67)$ & $98.08(\pm0.34)$ & $86.34(\pm0.72)$ \\
		SR & $81.10(\pm3.31)$ & $66.43(\pm2.39)$ & $69.80(\pm1.99)$ & $70.39(\pm1.67)$ & $87.24(\pm0.84)$ & $96.55(\pm1.14)$ & $77.56(\pm1.09)$ \\
		N2V & $81.47(\pm4.48)$ & $77.98(\pm1.54)$ & $82.53(\pm1.12)$ & $77.01(\pm1.79)$ & $83.08(\pm1.36)$ & $96.81(\pm0.86)$ & $78.48(\pm1.03)$ \\
		SEAL & $93.58(\pm0.68)$ & $86.49(\pm1.08)$ & $95.46(\pm0.59)$ & $91.90(\pm0.31)$ & $91.93(\pm0.31)$ & $99.51(\pm0.01)$ &$91.85(\pm0.20)$ \\ 
		LGLP & \textbf{95.46$(\pm0.43)$} & \textbf{89.70$(\pm0.53)$} & \textbf{97.37$(\pm0.25)$} & \textbf{92.92$(\pm0.21)$} & \textbf{92.61$(\pm0.23)$} & \textbf{99.82$(\pm0.01)$} & \textbf{92.98$(\pm0.10)$}\\
		\hline  
		\hline
		Model & Power & KHN & ADV & LDG & HPD & GRQ & ZWL\\
		\hline
		Katz & $74.29(\pm0.83)$ & $88.27(\pm0.32)$ & $93.72(\pm0.16)$ & $94.91(\pm0.27)$ & $89.52(\pm0.32)$ & $93.08(\pm0.29)$ & $97.08(\pm0.09)$ \\
		PR & $74.74(\pm0.81)$ & $92.17(\pm0.24)$ & $94.03(\pm0.24)$ & $96.26(\pm0.22)$ & $91.01(\pm0.23)$ & $93.18(\pm0.34)$ & $97.69(\pm0.08)$ \\
		SR & $70.69(\pm0.67)$ & $77.16(\pm0.81)$ & $83.31(\pm0.35)$ & $88.71(\pm0.79)$ & $84.16(\pm0.42)$ & $92.97(\pm0.31)$ & $95.44(\pm0.15)$ \\
		N2V & $76.55(\pm0.75)$ & $83.26(\pm0.79)$ & $79.02(\pm0.65)$ & $92.12(\pm0.50)$ & $80.57(\pm0.81)$ & $93.92(\pm0.31)$ & $93.82(\pm0.39)$ \\
		SEAL & $83.91(\pm0.83)$ & $93.40(\pm0.13)$ & $95.18(\pm0.12)$ & $96.55(\pm0.11)$ & $93.41(\pm0.09)$ & $97.86(\pm0.11)$ & $97.54(\pm0.02)$ \\ 
		LGLP & \textbf{84.78$(\pm0.53)$} & \textbf{94.14$(\pm0.09)$} & \textbf{95.72$(\pm0.08)$} & \textbf{96.86$(\pm0.06)$} & \textbf{93.65$(\pm0.08)$} & \textbf{98.14$(\pm0.10)$} & \textbf{97.91$(\pm0.01)$}\\
		\hline  
	\end{tabular}
\end{table*}

In this work, we compare our proposed method with three high-order heuristic methods including Katz \cite{katz1953new}, PageRank (PR) \cite{brin2012reprint}, SimRank (SR) \cite{jeh2002simrank}. In addition, graph embedding method node2vec (N2V) \cite{grover2016node2vec} and the state-of-the-art method SEAL \cite{zhang2018link} are selected as baseline methods. 

\subsection{Experimental Setup }
To assess the performance of our proposed method, we randomly select 50\% of existing links as positive training samples, and the rest are used as positive test samples. In addition, the same number of non-existed links are randomly selected from the graph as negative samples for training and testing. To demonstrate the effectiveness of our proposed method with a different number of training samples, we also select 80\% training links for the experiments. 

The parameters of baseline methods are tuned to achieve the best performance on datasets. The damping factor in Katz method is set to 0.001. The damping factor in PageRank is set to 0.85. The constant factor in the SimRank is set to 0.8. The dimension of node embedding for node2vec is set to 128. 

For the SEAL framework, we employ the same setting as the original paper \cite{zhang2018link}. The $2-hop$ enclosing subgraph is extracted for the SEAL framework,  and the labeling function is the same as equation (\ref{eq:label}) in this work. Three graph convolution layers are employed to compute node embeddings, and the sort pooling \cite{zhang2018end} is used to generate a fixed-size feature vector for the enclosing subgraph. The output feature map for three graph convolution layers is set to 32. The ratio of the sort pooling layer is set to 0.6. Two 1-D convolution layers with the number of output channels as 16 and 32, and two fully connected layers are employed as a classifier to predict the existence of a link. The SEAL model is trained for 50 epochs on each dataset.

To guarantee the comparison between our proposed method and SEAL model is fair, we employ the same graph neural network architecture to compute node embeddings in the line graph. It is worth noting that our proposed method does not employ graph pooling and 1-D convolution layers. Therefore, the number of parameters in our proposed method is much fewer than that in the SEAL model. Our proposed method is trained for 15 epochs on each dataset.

\begin{table*}[t]
	\caption{AUC comparison with baseline methods (50\% training links).}
	\label{result501}
	\centering
	\begin{tabular}{lccccccc}
		\hline
		Model & BUP & C.ele & USAir & SMG & EML & NSC & YST\\
		\hline
		Katz & $81.61(\pm3.40)$ & $79.99(\pm0.59)$ & $88.91(\pm0.39)$ & $80.65(\pm0.58)$ & $84.16(\pm0.64)$ & $95.99(\pm0.62)$ & $77.28(\pm0.37)$ \\
		PR & $84.07(\pm3.39)$ & \textbf{84.95$(\pm0.58)$} & $90.57(\pm0.39)$ & $84.59(\pm0.45)$ & $85.43(\pm0.63)$ & $96.06(\pm0.60)$ & $77.90(\pm3.69)$ \\
		SR & $80.98(\pm3.03)$ & $76.05(\pm0.80)$ & $81.09(\pm0.59)$ & $75.28(\pm0.74)$ & $83.05(\pm0.64)$ & $95.59(\pm0.68)$ & $73.71(\pm0.41)$ \\
		N2V & $80.94(\pm2.65)$ & $75.53(\pm1.23)$ & $84.63(\pm1.58)$ & $73.50(\pm1.22)$ & $80.15(\pm1.26)$ & $94.20(\pm1.25)$ & $73.62(\pm0.74)$ \\
		SEAL & $85.10(\pm0.82)$ & $81.23(\pm1.52)$ & $93.23(\pm1.46)$ & $86.56(\pm0.53)$ & $85.83(\pm0.46)$ & $99.07(\pm0.02)$ &$85.56(\pm0.28)$ \\ 
		LGLP & \textbf{88.57$(\pm0.52)$} & 84.60$(\pm0.82)$ & \textbf{95.18$(\pm0.33)$} & \textbf{89.54$(\pm0.36)$} & \textbf{86.77$(\pm0.26)$} & \textbf{99.33$(\pm0.01)$} & \textbf{87.63$(\pm0.15)$}\\
		\hline  
		\hline
		Model & Power & KHN & ADV & LDG & HPD & GRQ & ZWL\\
		\hline
		Katz & $57.34(\pm0.51)$ & $78.99(\pm0.20)$ & $90.04(\pm0.17)$ & $88.61(\pm0.19)$ & $81.60(\pm0.12)$ & $82.50(\pm0.21)$ & $93.72(\pm0.06)$ \\
		PR & $57.34(\pm0.52)$ & $82.34(\pm0.21)$ & $90.97(\pm0.15)$ & $90.50(\pm0.19)$ & $83.15(\pm0.17)$ & $82.64(\pm0.22)$ & 95.11$(\pm0.09)$ \\
		SR & $56.16(\pm0.45)$ & $75.87(\pm0.19)$ & $84.87(\pm0.14)$ & $87.95(\pm0.14)$ & $78.88(\pm0.22)$ & $82.68(\pm0.24)$ & $94.00(\pm0.10)$ \\
		N2V & $55.40(\pm0.84)$ & $78.53(\pm0.72)$ & $74.67(\pm0.98)$ & $88.82(\pm0.44)$ & $75.84(\pm1.03)$ & $84.24(\pm0.35)$ & $92.06(\pm0.61)$ \\
		SEAL & $65.80(\pm1.10)$ & $87.43(\pm0.17)$ & $92.75(\pm0.14)$ & $92.98(\pm0.16)$ & $88.05(\pm0.10)$ & $90.07(\pm0.15)$ & $94.94(\pm0.02)$ \\ 
		LGLP & \textbf{66.94$(\pm0.60)$} & \textbf{88.88$(\pm0.13)$} & \textbf{93.28$(\pm0.10)$} & \textbf{93.43$(\pm0.11)$} & \textbf{88.65$(\pm0.09)$} & \textbf{91.31$(\pm0.11)$} & \textbf{95.51$(\pm0.01)$} \\
		\hline  
	\end{tabular}
\end{table*}

\begin{table*}[t]
	\caption{AP comparison with baseline methods (50\% training links).}
	\label{result502}
	\centering
	\begin{tabular}{lccccccc}
		\hline
		Model & BUP & C.ele & USAir & SMG & EML & NSC & YST\\
		\hline
		Katz & $85.94(\pm2.03)$ & $83.99(\pm0.79)$ & $93.51(\pm0.35)$ & $87.68(\pm0.79)$ & $80.54(\pm0.31)$ & $98.02(\pm0.53)$ & $81.63(\pm0.41)$ \\
		PR & $89.53(\pm2.58)$ & \textbf{87.96$(\pm0.86)$} & $94.30(\pm0.49)$ & $91.07(\pm0.69)$ & $91.01(\pm0.52)$ & $98.08(\pm0.59)$ & $82.08(\pm0.46)$ \\
		SR & $81.09(\pm2.57)$ & $66.43(\pm1.17)$ & $69.78(\pm0.84)$ & $70.39(\pm0.96)$ & $87.24(\pm0.52)$ & $96.55(\pm0.75)$ & $76.02(\pm0.49)$ \\
		N2V & $76.05(\pm3.20)$ & $73.37(\pm1.23)$ & $81.03(\pm1.18)$ & $73.32(\pm1.34)$ & $81.12(\pm0.92)$ & $95.32(\pm1.08)$ & $76.61(\pm0.94)$ \\
		SEAL & $84.17(\pm0.62)$ & $83.94(\pm1.31)$ & $94.31(\pm1.13)$ & $86.76(\pm0.41)$ & $87.45(\pm0.41)$ & $99.09(\pm0.02)$ &$86.45(\pm0.25)$ \\ 
		LGLP & \textbf{89.03$(\pm0.41)$} & 84.80$(\pm0.63)$ & \textbf{94.89$(\pm0.33)$} & \textbf{90.23$(\pm0.26)$} & \textbf{88.49$(\pm0.23)$} & \textbf{99.38$(\pm0.01)$} & \textbf{89.22$(\pm0.13)$}\\
		\hline  
		\hline
		Model & Power & KHN & ADV & LDG & HPD & GRQ & ZWL\\
		\hline
		Katz & $57.63(\pm0.51)$ & $83.04(\pm0.38)$ & $91.76(\pm0.15)$ & $91.57(\pm0.17)$ & $85.73(\pm0.89)$ & $86.59(\pm0.20)$ & $95.12(\pm0.05)$ \\
		PR & $57.61(\pm0.56)$ & $87.18(\pm0.26)$ & $92.43(\pm0.17)$ & $93.53(\pm0.14)$ & $87.20(\pm0.15)$ & $86.73(\pm0.20)$ & 96.24$(\pm0.05)$ \\
		SR & $56.19(\pm0.49)$ & $75.87(\pm0.66)$ & $83.22(\pm0.20)$ & $88.11(\pm0.25)$ & $81.07(\pm0.18)$ & $86.27(\pm0.20)$ & $94.26(\pm0.11)$ \\
		N2V & $60.46(\pm0.86)$ & $80.60(\pm0.74)$ & $76.70(\pm0.82)$ & $89.57(\pm0.64)$ & $77.66(\pm0.54)$ & $88.70(\pm0.26)$ & $91.61(\pm0.49)$ \\
		SEAL & $68.67(\pm0.98)$ & $90.37(\pm0.16)$ & $93.52(\pm0.13)$ & $94.33(\pm0.15)$ & $90.25(\pm0.10)$ & $92.80(\pm0.12)$ & $95.88(\pm0.02)$ \\ 
		LGLP & \textbf{69.41$(\pm0.50)$} & \textbf{90.83$(\pm0.11)$} & \textbf{93.82$(\pm0.10)$} & \textbf{94.63$(\pm0.10)$} & \textbf{90.34$(\pm0.09)$} & \textbf{93.01$(\pm0.10)$} & \textbf{96.19$(\pm0.01)$} \\
		\hline  
	\end{tabular}
\end{table*}

\begin{figure*}[t]
	\center
	\includegraphics[width=\textwidth]{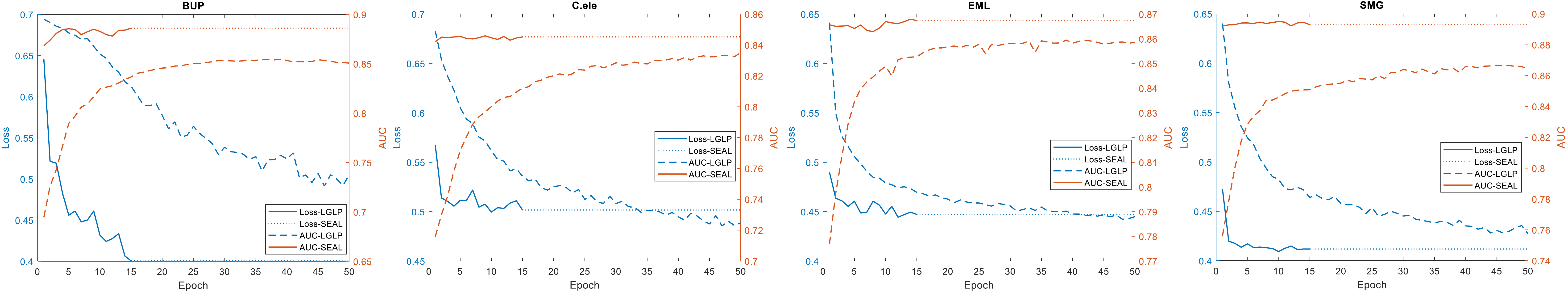}
	\caption{Training loss and testing AUC comparison between our proposed LGLP and SEAL method. The training loss and testing AUC on BUP, C.ele, EML, and SMG dataset. The training loss and testing AUC of LGLP are marked with blue and orange solid lines. Those of SEAL are marked with blue, orange dashed lines.
	} \label{fig:converage}
\end{figure*}

\begin{figure*}[t]
	\center
	\includegraphics[width=\textwidth]{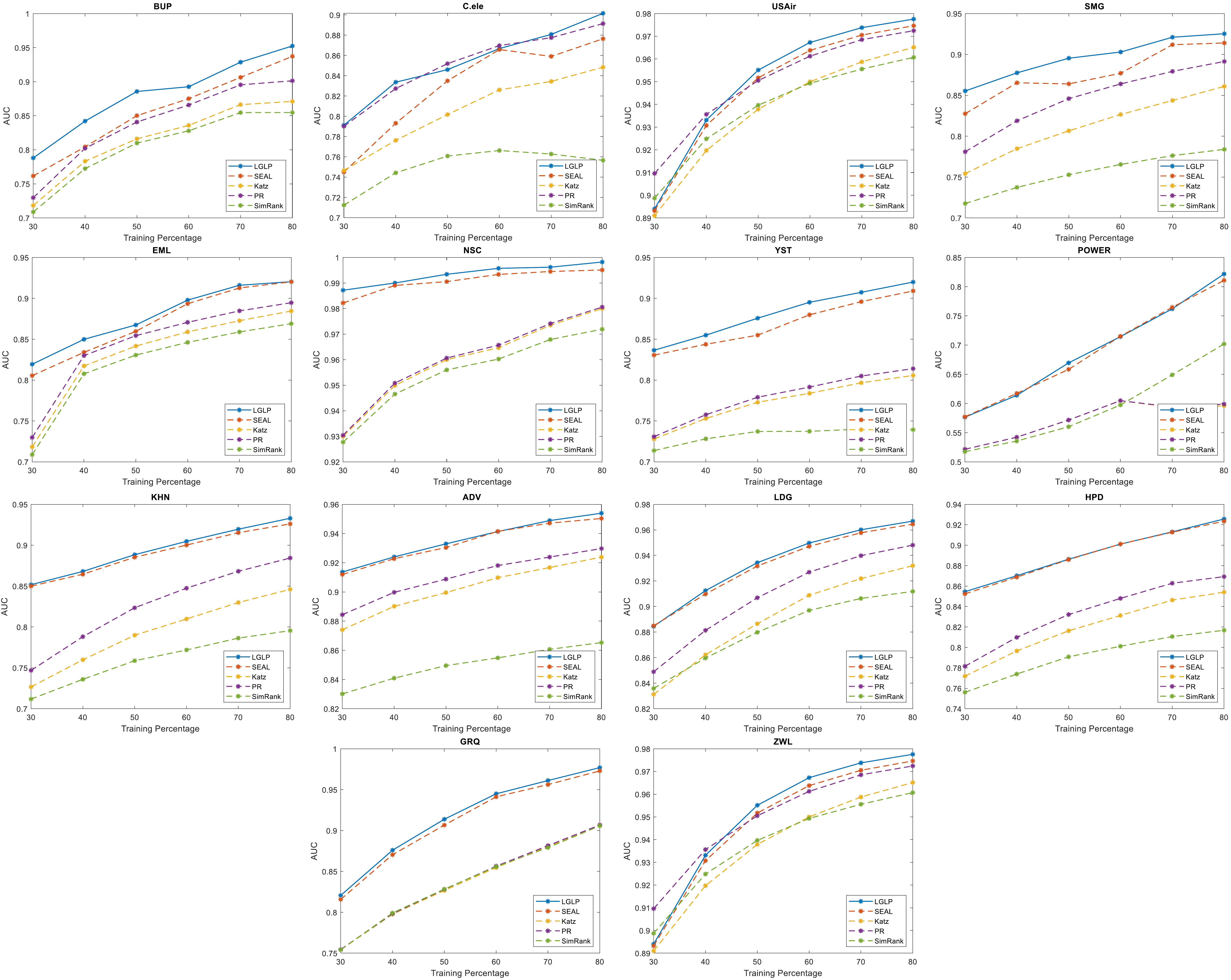}
	\caption{AUC comparison on all datasets for Katz, PR, SR, SEAL, and LGLP using different percent of training links. On each dataset, we take 30\%, 40\%, 50\%, 60\%, 70\%, and 80\% of all the links in $G$ as the training set. Our proposed method LGLP is marked with solid line and all baseline methods are marked with dashed lines in different colors.
	} \label{fig:percent}
\end{figure*}

\subsection{Results and Analysis}
\textbf{Plain Graph Link Prediction} We perform our proposed method and baseline methods on 14 datasets to compare the performance of each model. We randomly split each dataset into training and testing dataset for ten times. The averaged AUC and standard deviations using 80\% training links are shown in Table \ref{result801}. The results in terms of AP are shown in Table \ref{result802}. It can be seen from results that heuristic methods cannot achieve satisfactory performance on all datasets since the heuristic function is manually designed thus cannot handle different cases. We find that the state-of-the-art model SEAL always outperforms all heuristic methods and embedding methods since it can learn the distribution of links automatically from datasets. Our proposed LGLP model can consistently achieve better performance than all baseline methods, including SEAL in terms of two evaluation metrics. It shows that our proposed method can learn better features to represent the target link for prediction in line graph space. In addition, our proposed method is more stable than other baseline methods.

To demonstrate our proposed method can still achieve satisfactory performance with limited training samples, we conduct experiments on all datasets using 50\% training links. The averaged AUC and AP are shown in Table \ref{result501} and Table \ref{result502}, respectively. It can be seen from the results that our proposed method outperforms all baseline methods significantly on most datasets. We find that our proposed method can still perform well, even using 50\% training links. The AUC and AP are close to that of using 80\% training links. 

\begin{table}[t]
	\caption{Comparison on Cora dataset using plain graph and attributed graph (50\% training links).}
	\centering
	\label{tbl: att}
	\resizebox{0.37\textwidth}{!}{
		\begin{tabular}{lcccc}
			\hline
			\multirow{2}{*}{} & \multicolumn{2}{c}{Attribute} & \multicolumn{2}{c}{Plain} \\ 
			& AUC            & AP            & AUC              & AP              \\ \hline \hline
			SEAL              & 75.33          & 77.69          & 79.95            & 82.91           \\ 
			LGLP              & \textbf{81.45}           & \textbf{81.99}          & \textbf{79.96}             & \textbf{83.30}            \\ \hline
		\end{tabular}
	}
\end{table}

In the experiments, we dynamically take 30\%, 40\%, 50\%, 60\%, 70\%, and 80\% of all the links in $G$ as the training set and the rest as the test set, respectively. We conduct experiments with different training percentages and describe the AUC results in Figure \ref{fig:percent}. The AUC value of our proposed method is marked with a sold line, and other baseline methods are marked with dashed lines in different colors. It can be seen from the results that our proposed method can outperform all baseline methods with different percentages of the training data. In addition, the performance of our proposed method is not sensitive to the number of training samples.

\textbf{Attributed Graph Link Prediction} We also conduct experiments on attributed graphs. Since the heuristic method can only be applied to plain graphs, we mainly focus on the comparison between our proposed method and SEAL. In the SEAL framework, the attribute is concatenated with the node label as the input for graph neural networks. In this work, we propose a new function to combine the node label and node attributes. We perform the experiment on Cora dataset \cite{vsubelj2013model} that contains 2,708 nodes and 5,429 links. Each node in the Cora dataset is associated with an attribute vector in 1433 dimensions. We conduct the experiments without node attribute first, and then involve the node attributes to analyze the performance. The results are shown in Table \ref{tbl: att}. We can find both AUC and AP decrease after using node attributes as input in the SEAL model. In our proposed method, the performance does not change significantly. It shows that our proposed method can work well for both plain and attributed graphs.

\textbf{Convergence Speed Analysis} Our proposed method can learn features for the target link directly in the line graph. Therefore, only graph convolution layers are required to extract features. In the SEAL model, the procedure is completed in the original graph and thus requires graph convolution and pooling layers to achieve this goal. Compared with the SEAL model, our proposed method contains fewer parameters and converges faster. To analyze the converging speed of two models, we run the models on different datasets and collect the loss and test AUC value for each epoch. The result is shown in Figure \ref{fig:converage}. The loss and AUC of our proposed method are marked with solid lines. Those of the SEAL model are marked with dashed lines. It can be seen from the results that our proposed model can converge faster than the SEAL. Only 10 to 15 epochs are required to achieve the best performance for our proposed method. It takes 50 epochs for SEAL to converge. Therefore, our proposed method saves training time and requires fewer model parameters.

\section{Conclusion}

In this work, we propose a novel link prediction model based on line graph neural networks. Graph neural networks have achieved promising performance for the link prediction task. To deal with graphs in different scales, graph pooling layers are employed to extract a fixed-size feature vector in predicting the existence of a link. However, valuable information can be ignored in the pooling operation. In addition, graph neural networks with pooling layers commonly require more training time to converge. To overcome these limitations, we propose to transform the original input graph into line graph and thus the feature of the target link can be learned directly in the line graph without pooling operation. Experimental results on 14 datasets from different areas demonstrate that our proposed method can outperform all baseline methods, including the state-of-the-art models. In addition, our proposed method can converge faster than the state-of-the-art model significantly. 

\section*{Acknowledgment}

This work was supported in part by National Science Foundation grant DBI-2028361.

\bibliography{deep}
\bibliographystyle{IEEEtran}

\end{document}